\documentclass[10pt,twocolumn,letterpaper]{article}

\usepackage{wacv}
\usepackage{times}
\usepackage{epsfig}
\usepackage{graphicx}
\usepackage{amsmath}
\usepackage{amssymb}

\usepackage{url}            
\usepackage{booktabs}       
\usepackage{amsfonts}       
\usepackage{nicefrac}       
\usepackage{microtype}      
\usepackage{enumitem}
\usepackage{subfig}
\usepackage{amsmath,amsfonts,bm}

\def\eqref#1{equation~\ref{#1}}

\def\1{\bm{1}}

\def\eps{{\epsilon}}

\DeclareMathAlphabet{\mathsfit}{\encodingdefault}{\sfdefault}{m}{sl}
\SetMathAlphabet{\mathsfit}{bold}{\encodingdefault}{\sfdefault}{bx}{n}

\DeclareMathOperator*{\argmin}{arg\,min}

\def\wacvPaperID{1251} 

\wacvfinalcopy 
\ifwacvfinal
\def\assignedStartPage{9876} 
\fi

\ifwacvfinal
\usepackage[breaklinks=true,bookmarks=false]{hyperref}
\else
\usepackage[pagebackref=true,breaklinks=true,colorlinks,bookmarks=false]{hyperref}
\fi

\ifwacvfinal
\else
\fi

\begin{document}

\title{Compositional Embeddings for Multi-Label One-Shot Learning}

\author{Zeqian Li\\
Worcester Polytechnic Institute \\
{\tt\small zli14@wpi.edu}
\and
Michael Mozer\\
Google Brain\\
{\tt\small mcmozer@google.com}
\and
Jacob Whitehill\\
Worcester Polytechnic Institute \\
{\tt\small jrwhitehill@wpi.edu}
}

\maketitle
\thispagestyle{empty}

\begin{abstract}
  We present a \emph{compositional embedding} framework that infers not just a single class per input image,
  but a set of classes, in the setting of one-shot learning. Specifically, 
  we propose and evaluate several novel models consisting of  (1) an embedding function $f$ trained jointly with a ``composition'' 
  function $g$ that computes \emph{set union} operations between the classes encoded in two embedding vectors; and 
  (2) embedding $f$ trained jointly with  a ``query'' function $h$ that computes whether the classes encoded in one embedding \emph{subsume} 
  the classes encoded in another embedding. In
  contrast to prior work, these models must both \emph{perceive} the classes associated with the input 
  examples and \emph{encode} the relationships between different class label sets, and they are trained using only weak one-shot supervision
  consisting of the label-set relationships among training examples.
  Experiments on the OmniGlot, Open Images, and COCO datasets show that the proposed compositional embedding models
  outperform existing embedding methods. Our compositional embedding models have
  applications to multi-label object recognition for both one-shot and supervised learning.
\end{abstract}

\section{Introduction}
Embeddings, especially as enabled by advances in deep learning, have found widespread use in
natural language processing, object recognition, face identification and verification,
speaker verification and diarization, i.e., who is speaking when \cite{sell2018diarization},
and other areas. What embedding functions have in common is that
they map their input into a fixed-length distributed representation (i.e., continuous space) that facilitates more efficient
and accurate \cite{scott2018adapted} downstream analysis  than simplistic representations such as one-of-$k$ (one-hot).
Moreover, they are amenable to one-shot and few-shot learning since the set of classes
that can be represented does not depend directly on the dimensionality of the embedding space.

\begin{figure}
\begin{center}
\includegraphics[width=2.5in]{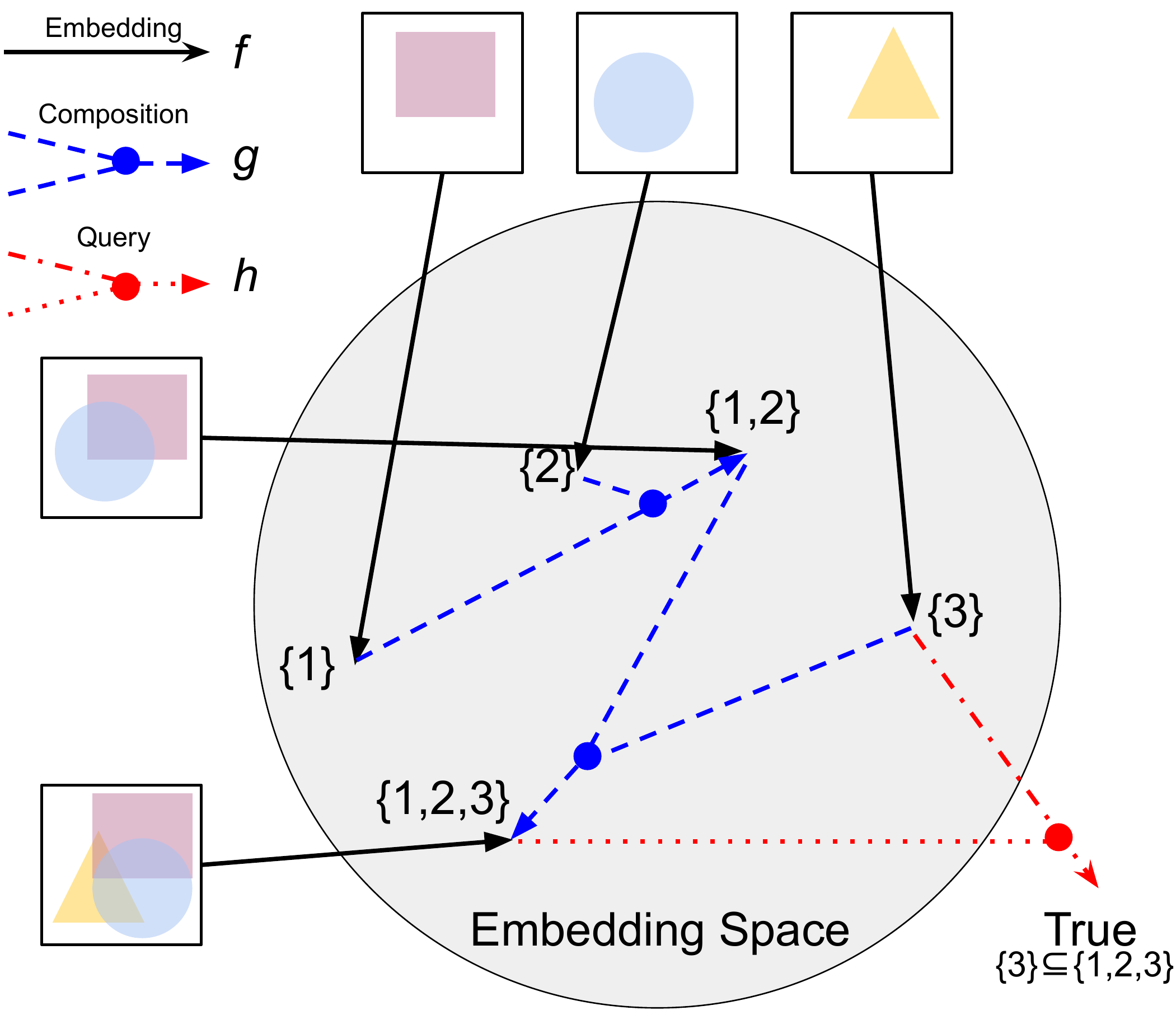}
\caption{
Overview: embedding function $f$ is trained jointly with either
a composition function $g$ or a query function $h$. In particular, $g$'s goal is to ``compose'' the embeddings of two examples,
containing classes $\mathcal{T}$ and $\mathcal{U}$ respectively, to approximate the embedding
of an example containing classes $\mathcal{T} \cup \mathcal{U}$.}
\label{fig:overview}
\end{center}
\end{figure}

The focus of most previous research on embeddings has been on cases where each example is associated with just one class (e.g., the image
contains only one person's face). In contrast, we investigate the case where each example 
is associated with not just one, but a \emph{subset} of classes from a universe $\mathcal{S}$.
Given 3 examples $x_a$, $x_b$ and $x_c$, the goal is to embed each example so that questions of two types can be answered (see Fig.~\ref{fig:overview}):
(1) Is the set of classes in example $x_a$ equal to the \emph{union} of the classes in examples $x_b$ and $x_c$? (2)
Does the set of classes in example $x_a$ \emph{subsume} the set of classes in example $x_b$?
For both these questions, we focus on settings in which the classes present in the example must be perceived automatically.

We approach this problem using \emph{compositional embeddings}. Like traditional embeddings, we train a function
$f$ that maps each example $x \in \mathbb{R}^n$ into an embedding space $\mathbb{R}^m$ so that examples with the same classes are mapped
close together and examples with different classes 
are mapped far apart. Unlike traditional embeddings, our function $f$ is trained to  represent the \emph{set} of classes that is
associated with each example, so that questions about set union and subsumption can be answered by comparing vectors in the embedding space.
We do not assume that the mechanism by which examples are rendered  from multiple classes is known.
Rather, the rendering process must be learned from training data.
We propose two models for one-shot learning, whereby $f$ is trained jointly with either a ``composition''
function $g$ that answers questions about set union, or a ``query'' function $h$ that answers question about subsumption (see
Figure \ref{fig:overview}).
This work has applications to multi-object recognition in images:
Given the embedding of an image $x_a$, answer whether $x_a$ contains the object(s) 
in another image $x_b$, where the latter could contain classes that were never observed
during training (i.e., one-shot learning).
Storing just the embeddings but not the pixels could be more space-efficient and provide
a form of image compression.

{\bf Contributions}: To our best knowledge, our model is the first to perform multi-class one-shot learning 
using only weak supervision consisting of label-set relationships between training
examples (in contrast to the strongly-supervised training approach in \cite{alfassy2019laso}; see the last section in Related Work).
We explore
how embedding functions can be trained both to \emph{perceive} multiple objects that are  possibly entangled (overlapping in space)
and to \emph{represent} them so that set operations can be conducted among embedded vectors.
We instantiate this idea in two ways -- Model I for set union ($f$\&$g$) and Model II for set containment ($f$\&$h$) --
and evaluate our models on visual domain. Our experiments show promising results that compositional embeddings
can perceive and compute set relationships in highly challenging perceptual contexts. Since one-shot learning for multi-label
classification is a new problem domain, we devise baseline methods based on traditional (non-compositional) embeddings, and our experiments
provide evidence that compositional embeddings offer significant accuracy advantages.
Finally, we explore applications of compositional embeddings to multi-label
image classification for supervised (not one-shot) learning settings (Model III).

Supp.~Material includes appendix and code is available in \footnote{\url{https://drive.google.com/drive/folders/1zjsK9DP3CUqwcVSNwDPshIxOV5hQwFxt?usp=sharing}}.

\section{Related Work}
{\bf Embeddings}:
We distinguish two types of embeddings: (1) ``Perceptual'' embeddings
such as for vision (Facenet \cite{schroff2015facenet}) and speech (x-vector \cite{snyder2018x}),
where each class (e.g., person whose voice was recorded
or face was photographed) may contain widely varying examples across emotion,
lighting, background noise, etc. (2) Item embeddings for words (word2vec \cite{mikolov2013efficient}, GloVe \cite{pennington2014glove}),
or for items \& users in recommendation systems \cite{dai2016deep}; here, 
each class contains only one exemplar by definition.
Within the former, the task of the embedding
function is to map examples from the same class close together and examples from
other classes far apart. This often requires deep, non-linear transformations to be successful.
With item embeddings, the class of each example does not need to be inferred;
instead, the goal is to give the embedded vectors geometric structure to reflect co-occurrence, similarity
in meaning, etc.

{\bf Compositional embeddings}:
Most prior work on compositionality in embedding models has focused on word embeddings 
\cite{pollack1989implications,nakov2016semeval,lake2017generalization,joshi2018pair2vec}.
More recent work has explored ``perceptual'' embeddings:
\cite{dey2018learning} combined embeddings from different sources in multimodal training. 
\cite{kulkarni20193d} proposed a model to infer the relative pose from embeddings of objects in the same scene.
\cite{tokmakov2019learning} proposed a method to decompose the attributes of one object into multiple representations.
\cite{andreas2019measuring} introduced a method for generating judgments about compositional structure in embeddings.
\cite{sylvain2019locality} showed that compositional information was deeply related to generalization in zero-shot learning.
\cite{stone2017teaching} proposed a way to train CNNs to learn features that had compositional property so that objects could be separated 
better from their surroundings and each other.
\cite{Lyu2019} used compositional network embeddings to predict whether two new nodes in a graph, which were not observed during training,
are adjacent, using node-based features as predictors.

{\bf Multi-label few-shot learning}:
The last few years have seen some emerging interest in the field of multi-label few-shot and zero-shot learning.
\cite{santa2018neural} proposed a network to learn basic visual concepts and used compositional concepts to represent singleton objects not seen during training. 
Liu et al.~\cite{liu2018temporal} introduced an approach to infer compositional language descriptions of video activities and achieved zero-shot learning by composing seen descriptions.
Huynh and Elhamifar \cite{huynh2020shared} proposed a visual attention-based method for multi-label image classification that can generalize to classes not seen during training, but it requires auxiliary semantic vectors (e.g., attributes or word embeddings) associated with the unseen classes.
The most similar work to ours is by Alfassy et al.~\cite{alfassy2019laso}. Their model also tackles the problem of
generating latent representations that reflect the set of labels associated with each input, and it also uses trained 
set operations (union, intersection, difference) that operate on pairs of examples.
Algorithmically, our work differs from Alfassy's in several ways:
Their method depends on strong supervision whereby the embedding and composition functions are trained using a fixed set of classes (they train on 64 classes from COCO), such that each image in the training set must be labeled w.r.t.~\emph{all} the classes in the \emph{entire} training  set.
Their method also requires an extra multi-label classifier, and as a result they have 4 separate losses 
that are applied at different points during training.  In contrast, our model requires only weak supervision: Each training episode has its own subset of classes, and each image in the episode must be labeled only w.r.t.~that subset -- there is no need for it to be labeled w.r.t.~all classes in the entire training set, or even for the set of training classes to be finite. Also, each of our models is trained using just 1 loss function.
To emphasize, their approach requires class-labeled data for a fixed set of classes, whereas our approach requires merely sets of examples that possess certain compositional relationships. 

\section{Model I: Embedding $f$  and Composition $g$}
{\bf Assumptions and notation}: 
For generality, we refer to the data to be embedded (images, videos, etc.) simply as ``examples''.
Let the universe of classes be $\mathcal{S}$.
From any subset $\mathcal{T} \subseteq \mathcal{S}$, a ground-truth rendering function
$r: 2^\mathcal{S} \rightarrow \mathbb{R}^n$ ``renders'' an example, i.e., $r(\mathcal{T})=x$.
Inversely, there is also a ground-truth classification function $c: \mathbb{R}^n \rightarrow 2^\mathcal{S}$ that
identifies the label set from the rendered example, i.e., $c(x)=\mathcal{T}$.  Neither $r$ nor $c$ is observed.
We let $e_\mathcal{T}$ represent the embedding (i.e., output of $f$) associated with some example containing classes $\mathcal{T}$.

\label{sec:model1}
{\bf Model}: Given two examples $x_a$ and $x_b$ that are associated with singleton sets $\{s\}$ and $\{t\}$, respectively, the hope is that, for some 
third  example $x_c$ associated with \emph{both} classes $\{s,t\}$, we have
\begin{equation}
g(f(x_a), f(x_b)) \approx f(x_c)
\label{eqn:gfunction}
\end{equation}
Moreover, we hope that $g$ can generalize to \emph{any} number of classes within the set $\mathcal{S}$. For example, if
example $x_d$ is associated with a singleton set $\{u\}$ and $x_e$ is an example associated with $\{s,t,u\}$, then we hope $g(g(f(x_a), f(x_b)), f(x_d)) \approx f(x_e)$.

There are two challenging tasks that $f$ and $g$ must solve cooperatively:
(1) $f$ has to learn to perceive multiple objects that appear simultaneously and are possibly non-linearly entangled with each other
-- all \emph{without} knowing the rendering process $r$ of how examples are formed or how classes are combined.
(2) $g$ has to define geometrical structure in the embedding space to support set unions.
One way to understand our computational problem is the following: If $f$ is invertible, then ideally we would want $g$ to compute
\begin{equation}
g(e_\mathcal{T},e_\mathcal{U})=f(r(c(f^{-1}(e_\mathcal{T})) \cup c(f^{-1}(e_\mathcal{U})))).
\label{eqn:gfunction2}
\end{equation}
In other words, one way that $g$ can perform well is to
learn (without knowing $r$ or $c$) to do the following:
(1) invert each of the two input embeddings; (2) classify the two corresponding label sets;
(3) render an example with the union of the two inferred label sets; and (4) embed the result.
Training $f$ and $g$ jointly may also ensure systematicity of the embedding space such that any combination of
objects can be embedded.

{\bf One-shot learning}:
Model I can be used for one-shot learning on a set of classes
$\mathcal{N} \subset \mathcal{S}$ not seen during training in the following way:
We obtain $k$ labeled examples $x_1,\ldots,x_k$ from the user, where each $\{s_i\} = c(x_i)$ is the singleton set formed from the
$i$th element of $\mathcal{N}$ and $|\mathcal{N}|=k$. We call these examples the \emph{reference examples}. (Note that $\mathcal{N}$ typically changes during each episode; hence, these reference examples provide only a weak form of supervision about the class labels.)
We then infer which set of classes is represented by a new example $x'$ using the following procedure:
(1) Compute the embedding of $x'$, i.e., $f(x')$.
(2) Use $f$ to compute the embedding of each singleton example $x_i$, i.e.,  $e_{\{i\}}=f(x_i)$.
(3) From $e_{\{1\}},\ldots,e_{\{k\}}$, estimate the embedding of \emph{every} subset $\mathcal{T}=\{s_1,\ldots,s_l\}\subseteq \mathcal{N}$ according
to the recurrence relation:
\begin{equation}
e_{\{s_1,\ldots,s_l\}} = g(e_{\{s_1,\ldots,s_{l-1}\}}, e_{\{s_l\}})
\label{eqn:g_recurrence}
\end{equation}
Finally, (4) estimate the label of $x'$ as 
\begin{equation}
\argmin_{\mathcal{T}\subseteq \mathcal{N}} |f(x') - e_\mathcal{T}|_2^2
\label{eqn:iterate}
\end{equation}

\subsection{Training Procedure}
\label{sec:training_f_g}
Functions $f$ and $g$ are trained jointly: For each example $x$ associated with classes $\mathcal{T}$, we compute $e_\mathcal{T}$ from the
singleton reference examples according to Eq.~\ref{eqn:g_recurrence}. (To decide the order in which we apply the recursion,
we define an arbitrary ordering over the elements of $\mathcal{N}$ and iterate accordingly.) We then compute 
a triplet loss
\begin{equation}
\max(0, ||f(x) - e_\mathcal{T}||_2 - ||f(x) - e_{\mathcal{T}'}||_2 + \eps)
\label{eqn:lossfunction}
\end{equation}
for every $\mathcal{T}' \ne \mathcal{T} \subseteq \mathcal{N}$, where $\epsilon$ is a small positive real number
\cite{weinberger2009distance,schroff2015facenet}.
In practice, for each example $x$, we randomly pick some
$\mathcal{T}'\in 2^\mathcal{N}$ for comparison. Both $f$ and $g$ are optimized jointly in backpropagation because the loss function is applied to embeddings generated from both.

Note that we also tried another method of  training $f$ and $g$ with the explicit goal of 
encouraging $g$ to map $e_\mathcal{T}$ and $e_\mathcal{U}$ to be close  
to $e_{\mathcal{T}\cup\mathcal{U}}$.  This can be done by training $f$ and $g$ alternately, or by training them 
jointly in the same backpropagation.  However, this approach yielded very poor results. A possible explanation 
is that $g$ could fulfill its goal by mapping all vectors to the same location (e.g., ${\bf 0}$). Hence, with this training method,
a trade-off
 arose between $g$'s goal and $f$'s goal (separating examples with distinct label sets). 


\subsection{Experiment 1: OmniGlot}
\label{sec:g_omniglot}
We first evaluated our method on 
the OmniGlot  dataset \cite{lake2015human}. OmniGlot contains handwritten characters from 50 different alphabets; in total it comprises
1623 symbols, each of which was drawn by 20 people and rendered as a 64$\times$64 image.
OmniGlot has been widely used in one-shot learning research
(e.g., \cite{rezende2016one,bertinetto2016learning}). 

In our experiment, the model is provided with one reference image for each singleton test class (5 classes in total). Then,
$f$ and $g$ are used to select the subset of classes that most closely match the embedding of each test example (Eq.~\ref{eqn:iterate}).
The goal is to train $f$ and $g$ so that, on classes not seen during training,
the exact set of classes contained in each test example can be inferred.

\begin{figure}
\begin{center}
\setlength{\tabcolsep}{1.5pt}
\begin{tabular}[b]{ccccc}
\includegraphics[width=0.45in]{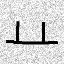} &
\includegraphics[width=0.45in]{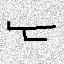} &
\includegraphics[width=0.45in]{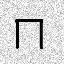} &
\includegraphics[width=0.45in]{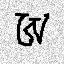} &
\includegraphics[width=0.45in]{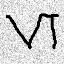} \\
 \{1\} & \{2\} & \{3\} & \{4\} & \{5\} \\ 
\includegraphics[width=0.45in]{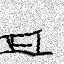} &
\includegraphics[width=0.45in]{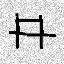} &
\includegraphics[width=0.45in]{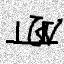} &
\includegraphics[width=0.45in]{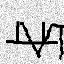} &
\includegraphics[width=0.45in]{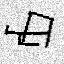} \\
 \{1,2\} & \{1,3\} & \{1,4\} & \{1,5\} & \{2,3\} \\ 
\includegraphics[width=0.45in]{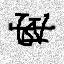} &
\includegraphics[width=0.45in]{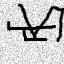} &
\includegraphics[width=0.45in]{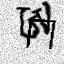} &
\includegraphics[width=0.45in]{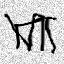} &
\includegraphics[width=0.45in]{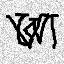} \\
 \{2,4\} & \{2,5\} & \{3,4\} & \{3,5\} & \{4,5\} \\ 
\includegraphics[width=0.45in]{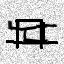} &
\includegraphics[width=0.45in]{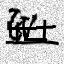} &
\includegraphics[width=0.45in]{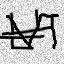} &
\includegraphics[width=0.45in]{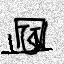} &
\includegraphics[width=0.45in]{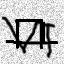} \\
 \{1,2,3\} & \{1,2,4\} & \{1,2,5\} & \{1,3,4\} & \{1,3,5\} \\ 
\includegraphics[width=0.45in]{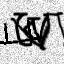} &
\includegraphics[width=0.45in]{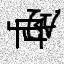} &
\includegraphics[width=0.45in]{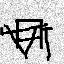} &
\includegraphics[width=0.45in]{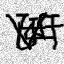} &
\includegraphics[width=0.45in]{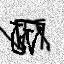} \\
 \{1,4,5\} & \{2,3,4\} & \{2,3,5\} & \{2,4,5\} & \{3,4,5\} \\ 
\end{tabular}
\end{center}
\caption{
Examples of the images 
from the OmniGlot dataset, used in Experiments 1 and 2.
Below each image is its associated class label set $\mathcal{T}$.
}
\label{fig:OmniExample}
\end{figure}

We assessed to what extent the proposed model can capture set union operations.
To create each example with label set $\mathcal{T}$, the rendering function $r$ randomly picks one of the 20 exemplars from each 
class $s\in\mathcal{T}$ and then randomly shifts, scales, and rotates it.
Then, $r$  computes the pixel-wise minimum across all the constituent images (one for each element of $\mathcal{T}$).
Finally, $r$ adds Gaussian noise.
See Fig.~\ref{fig:OmniExample} and Supp.~Material.  Due to the complexity of each character and the
overlapping pen strokes in composite images, recognizing the class label sets is challenging even for humans, especially for $|\mathcal{T}|=3$.

In this experiment, we let the total number of possible symbols in each episode be $k=5$. We trained $f$\&$g$ such that
the maximum class label set size was 3 (i.e., $|\mathcal{T}|\leq 3$). 
There are 25 such (non-empty) sets in total
(5 singletons, ${5\choose 2}=10$ 2-sets, and ${5\choose 3}=10$ 3-sets).

{\bf Architecture}:
For $f$, we used ResNet-18 \cite{he2016deep} that was modified to have 1 input channel and a 32-dimensional output.
For $g$, we tested several architectures.
First, define
Symm$(a,b;k)=W_1 a + W_1 b + W_2 (a\odot b)$ to be a symmetric function\footnote{
	To be completely order-agnostic, $g$ would have to be both symmetric and associative. Symmetry alone does not ensure $g(g(x,y), z) = g(x, g(y, z))$, but it provides at least some (if imperfect) invariance to order.} (with parameter matrices $W_1,W_2$) of its two examples $a,b \in \mathbb{R}^n$ that produces a vector in $\mathbb{R}^k$. We then defined four possible architectures for $g$:

\begin{itemize}[leftmargin=0.5cm]
\item {\bf Mean} ($g_\textrm{Mean}$): $\frac{(a+b)}{2}\rightarrow\textrm{L2Norm}$.

\item {\bf Bi-linear} ($g_\textrm{Lin}$): $\textrm{Symm}(a,b;32)\rightarrow\textrm{L2Norm}$.

\item {\bf Bi-linear + FC} ($g_\textrm{Lin+FC}$): $\textrm{Symm}(a,b;32)\rightarrow\textrm{BN}\rightarrow\textrm{ReLU}\rightarrow\textrm{FC}(32)\rightarrow\textrm{L2Norm}$.

\item {\bf DNN} ($g_\textrm{DNN}$): $
\textrm{Symm}(a,b;32)\rightarrow\textrm{BN}\rightarrow\textrm{ReLU}\rightarrow\textrm{FC}(32)\rightarrow
\textrm{BN}\rightarrow\textrm{ReLU}\rightarrow\textrm{FC}(32)\rightarrow\textrm{BN}\rightarrow\textrm{ReLU}\rightarrow
\textrm{FC}(32)\rightarrow\textrm{L2Norm}
$.
\end{itemize}
$\textrm{BN}$ is batch normalization, and 
FC$(n)$ is a fully-connected layer with $n$ neurons. 
We note that $g_\textrm{Mean}$ is similar to the implicit compositionality  found in
word embedding models \cite{mikolov2013efficient}.


{\bf Training}: For each mini-batch, $\mathcal{N}$ was created by randomly choosing 5 classes from the universe $\mathcal{S}$ (where $|\mathcal{S}|=944$ in training set).
Images from these classes are rendered using function $r$ from either singleton, 2-set class label sets, or
3-set class label sets. In other words, $1\leq |\mathcal{T}| \leq 3$ for all examples.
See Supp.~Material for details.

{\bf Testing}:
Testing data are generated similar to training data, but none of the classes were seen during training.
We optimize Eq.~\ref{eqn:iterate} to estimate the label set for each test example.


{\bf Baselines}: 
Because multi-label few-shot learning is a new learning domain, and because none of the existing literature exactly
matches the assumptions of our model (\cite{alfassy2019laso} assumes strongly supervised training labels, and
\cite{huynh2020shared} requires auxiliary semantic labels for unseen classes), it was not obvious to what baselines we should compare.
When evaluating our models, we sought to assess the unique contribution of the \emph{compositional} embedding above and beyond
what traditional embedding methods achieve. 
We compared to two baselines:
\begin{enumerate}[leftmargin=0.5cm]

\item {\bf Traditional embedding $f$ and average ({\bf TradEm})}: 
A reasonable hypothesis is that a traditional embedding function for one-shot learning
trained on images with singleton class label sets can implicitly generalize to
larger label sets by interpolating among the embedded vectors. Hence, 
we trained a traditional (i.e., non-compositional) embedding $f$ just on
singletons using one-shot learning, similar to \cite{ye2018deep,koch2015siamese}.
(Accuracy on singletons after training on OmniGlot: $97.9\%$ top-1 accuracy in classifying test examples over 5 classes.)
The embedding of a composite image with label set $\mathcal{T}$ is then estimated using the mean of the embeddings
of each class in $\mathcal{T}$. In contrast to $g_\textrm{Mean}$ above, the $f$ in this baseline is trained
by itself, without knowledge of how its embeddings will be composed.

{\bf Note}: In our experiments, the models always needed to pick the correct answer from 25 candidates. ``1-sets'' in the table means the accuracy when the ground truth is a singleton, but the model still sees 25 candidates. 


\item {\bf Most frequent ({\bf MF})}: Always guess the most frequent element in the test set. Since all classes occurred equally 
frequently, this was equivalent to random guessing. While simplistic, this baseline is useful to get a basic sense of how difficult the task is.

\end{enumerate}

{\bf Assessment}:
We assessed accuracy (\%-correct) in 3 ways:
(a) Accuracy, over all test examples, of identifying $\mathcal{T}$.
(b) Accuracy, over test examples for which $|\mathcal{T}|=l$ (where $l\in\{1,2,3\}$), of identifying $\mathcal{T}$.
{\bf Note}: we did \emph{not} give the models the benefit of knowing $|\mathcal{T}|$ -- each model predicted
the class label set over \emph{all} $\mathcal{T} \subset \mathcal{N}$ such that $|\mathcal{T}|\leq 3$.
This can reveal whether a model is more accurate on examples with fewer vs.~more classes.
(c) Accuracy, over all examples, in determining just the \emph{number} of classes in the set, i.e., $|\mathcal{T}|$.

{\bf Results}:
As shown in Table~\ref{fig:OmniResult}, the MF baseline accuracy 
was just $4\%$ for an exact (top-1) and $12\%$ for top-3 match.
(Recall that the models did not ``know'' $|\mathcal{T}|$ and 
needed to pick the correct answer from all 25 possible label sets.)
Using the TradEm approach,
accuracy increased to $25.5\%$ and $40.9\%$, respectively. All of the proposed 
$f$\&$g$ models strongly outperformed the TradEm baseline, indicating
that training $f$ jointly with a composition function is helpful.
For all the $f$\&$g$ approaches as well as the TradEm baseline,
model predictions were well above chance  (MF) for all label set sizes, i.e., these approaches could all distinguish
label sets with more than one element at least to some degree.

In terms of architecture for composition function $g$,
overall, the $g_\textrm{Lin}$, which contains a symmetric bi-linear layer before $L_2$-normalization, did best: $64.7\%$
and $87.6\%$ for top-1 and top-3 matches over all examples, respectively.
This  suggests that
composition by averaging alone is not optimal for this task.
However, adding more layers (i.e., $g_\textrm{Lin+FC}, g_\textrm{DNN}$) did not help, 
especially when $|\mathcal{T}|$ increases.
It is possible that the more complex $g$ 
overfit, and that with regularization or more training data the deeper models might prevail.



\begin{table}
\begin{center}
\setlength{\tabcolsep}{1.6pt}
\begin{tabular}[b]{l|l|c|c|c|c||c|c}
\multicolumn{8}{c}{\bf Experiment 1 (OmniGlot): Train with $|\mathcal{T}|\leq 3$ } \\\hline \hline
\multicolumn{8}{c}{\bf Label Set Identification} \\\hline
\multicolumn{2}{c|}{}& \multicolumn{4}{c||}{$f\& g$ Approaches} & \multicolumn{2}{c}{Baselines} \\\hline
&                            & $g_\textrm{DNN}$  & $g_\textrm{Lin+FC}$ & $g_\textrm{Lin}$ & $g_\textrm{Mean}$ & TradEm & MF \\ \hline
All    & Exact  & 50.6 & 56.7   & {\bf 64.7} & 52.8 & 25.5 &  4.0 \\ 
        & Top-3 & 76.5 & 81.7   & {\bf 87.6} & 80.0 & 40.9 & 12.0 \\   \hline
1-sets & Exact  & 94.5 & 96.0   & {\bf 97.0} & 86.9 & 89.3 & 4.0 \\  
        & Top-3 & 99.1 & 99.4   & {\bf 99.6} & 95.4 & 96.6 & 12.0 \\   \hline
2-sets & Exact  & 51.2 & 54.6   & {\bf 64.5} & 49.7 & 15.4 & 4.0 \\  
        & Top-3 & 82.9 & 83.0   & {\bf 87.9} & 81.4 & 37.7 & 12.0 \\   \hline
3-sets & Exact  & 27.9 & 39.1   & {\bf 48.9} & 39.0 & 3.7 & 4.0 \\  
        & Top-3 & 58.7 & 71.6   & {\bf 81.1} & 71.1 & 16.4 & 12.0 \\   \hline \hline
\multicolumn{8}{c}{\bf Set Size Determination} \\\hline
\multicolumn{2}{c|}{All}       & 81.7 & 87.4 & {\bf 90.1} & 71.4 & 44.9 & 36.0
\end{tabular}
\end{center}
\caption{
Experiment 1 (OmniGlot):
One-shot mean accuracy (\% correct) of Model I in inferring the label set of each example exactly (top 1), within the top 3, and the
size of each label set. Set Size Determination measures the ability to infer the set size. 
TradEm is similar to \cite{ye2018deep,koch2015siamese}, and MF is based on random guessing.
}
\label{fig:OmniResult}
\end{table}

{\bf Discussion}:
Experiments 1 suggests that, for $f$\&$g$ compositionality for set union, a simple linear layer works best.
Function $g_\textrm{Lin}$, despite the $\textrm{L2Norm}$ at the end,
might retain a greater degree of associativity (i.e., $(a+b)+c=a+(b+c)$) than deeper $g$ functions.
This property may be important especially for larger $\mathcal{T}$, where $g$ is invoked multiple times to create
larger and larger set unions.

{\bf Scalability}: The number of subsets is exponential in $|\mathcal{N}|$, which poses a scalability problem for both training and testing,
and hence Model I may in some sense be regarded more as a proof-of-concept than practical algorithm. However,
in settings where the number of simultaneously present classes is inherently small (e.g., in speaker diarization from audio signals, it is rare for 
more than just a few people to speak at once), the model can still be practical. In our Model II (Section \ref{sec:model2}),
we overcome this scalability issue by switching from set union to set containment.

\section{Model II: Embedding $f$ and Query $h$}
\label{sec:model2}
With this model we explore compositional embeddings that implements \emph{set containment}:
In some applications, it may be more useful to determine whether an example \emph{contains} an object or set of objects.
For instance, we might want to know whether a specific object is contained in an image.
Moreover, in some settings, it may be difficult during training to label every example (image, video, etc.) for the presence of
\emph{all} the objects it contains -- for each example, we might only know its labels for a subset of classes.
Here we propose a second type of compositional embedding mechanism that tests whether the set of classes associated  with one
example \emph{subsumes} the set of classes associated with another example. We implement this using a ``query'' function $h$ that takes two 
embedded examples as inputs:
$h(f(x_a), f(x_b)) = \textrm{True} \iff c(x_b) \subseteq c(x_a)$.
Note that $h$ can be trained with only weak supervision w.r.t.~the individual examples: it only needs to know 
\emph{pairwise} information about which examples ``subsume'' other examples.
Compared with typical multiple instance learning models, Model II deals with single samples instead of bags of instances. Additionally, training procedure of Model II is more focused on one-shot learning.


\subsection{Training procedure}
Functions $f$ and $h$ are trained jointly.
Since $h$ is not symmetric, its first layer is replaced with a linear layer $W_1 a + W_2 b$ (see Supp.~Material).
In contrast to Model I, reference examples are not needed; only the subset relationships
between label sets of pairs of examples are required.
We backpropagate a binary cross-entropy loss, based on correctly answering the query defined above,
through $h$ to $f$.

\subsection{Experiment 2: OmniGlot}
\begin{table}
\begin{center}
\setlength{\tabcolsep}{1.6pt}
\begin{tabular}{l||c|c|c||c}
\multicolumn{5}{c}{\bf Experiment 2 (OmniGlot)} \\\hline \hline
  & $h_\textrm{DNN}$ & $h_\textrm{Lin+FC}$ & $h_\textrm{Lin}$ & TradEm  \\ \hline
Acc \% & {\bf 71.8} & 71.1 & 50.8 & 63.8 \\ \hline
AUC & {\bf 80.0} & 79.1 & 51.4 & 78.2
\end{tabular}
\caption{One-shot learning results for Model II (with different versions of $h$) on OmniGlot compared to a traditional
(non-compositional) embedding baseline (TradEm).
}
\label{fig:query_omniglot}
\end{center}
\end{table}

Here we assess Model II on OmniGlot where size of class label sets is up to 5, and we use the same rendering function $r$ in Experiment 1.
Let $f(x_a)$ and $f(x_b)$ be the two arguments to $h$. For $x_a$, each image can be associated with multiple classes, from 1 class (i.e.,
$c(x_a) = \{s_1\}$) to 5 classes (i.e., $c(x_a) = \{s_1, s_2, \ldots, s_5\}$), where all label sets occur with equal frequency.
For $x_b$ (which is always a singleton in this experiment), half are positive examples (i.e., such that $h(f(x_a), f(x_b))=\textrm{True}$) 
which are associated with classes contained in $x_a$, so that $c(x_b) \subseteq c(x_a)$. The other half are negative examples ($h(f(x_a), f(x_b))=\textrm{False}$), 
where $x_b$ is associated with some other singleton class $c(x_b) \not\subseteq c(x_a)$. Both the training set and test set have this configuration.

{\bf Architecture}: 
The $f$ was the same as in Experiment 1.
For $h$, we tried several functions ($h_\textrm{DNN},h_\textrm{Lin+FC},h_\textrm{Lin}$),
analogous to the different $g$ from Section \ref{sec:g_omniglot} except the final layers
are 1-dim sigmoids.
See Supp.~Materials.

{\bf Baseline}: How would we tackle this problem without compositional embeddings?
We compared our method with a traditional (non-compositional) embedding method ({\bf TradEm}) that is trained to separate
examples according to their association with just a \emph{single} class. In particular, for each composite example $x_a$ (i.e., $|c(x_a)|=2$),
we picked one of the two classes arbitrarily (according to some fixed ordering on the elements of $\mathcal{S}$); call this class $s_1$.
Then, we chose a positive example $x_b$ (such that $c(x_b)=\{s_1\}$) and a negative example $x_c$ (such that $c(x_c) = \{s_3\} \not \subseteq c(x_a)$).
We then compute a triplet loss so the distance between $f(x_a)$ and $f(x_b)$ is smaller than the distance between $f(x_a)$ and $f(x_c)$,
and backpropagate the loss through $f$. During testing,  we use $f$ to answer a query---does $c(x_a)$ contain $c(x_b)$?---%
by thresholding ($0.5$) the distance between $f(x_a)$ and $f(x_b)$.

{\bf Results} are shown in Table~\ref{fig:query_omniglot}. Compositional embeddings, as implemented with a combination of $f$
trained jointly with either $h_\textrm{DNN}$ or $h_\textrm{Lin+FC}$, outperform the TradEm baseline,
in terms of both \% correct accuracy and AUC. Unlike in Model I, where $h_\textrm{Lin}$ achieved the best results, $f$ trained jointly with
$h_\textrm{Lin}$ is just slightly better than random guess (50\%). The deeper $h$ worked better.

\subsection{Experiment 3: Open Images}
\begin{figure}
\begin{center}
\includegraphics[width=1.5in]{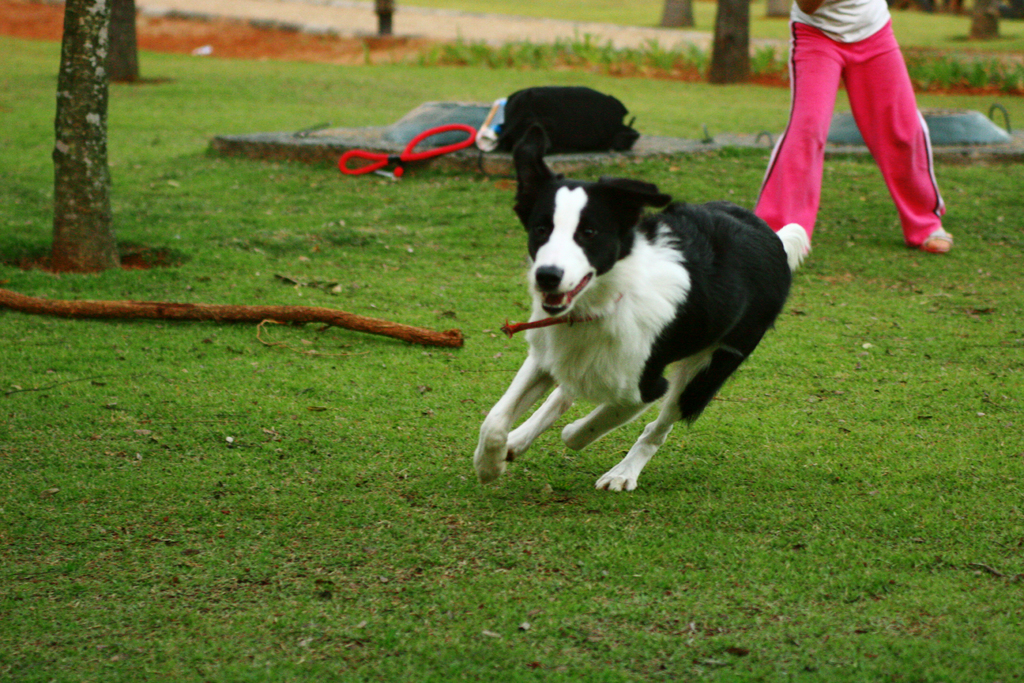}
\setlength{\tabcolsep}{0.6pt}
\begin{tabular}[b]{ccccc}
\includegraphics[width=.625in]{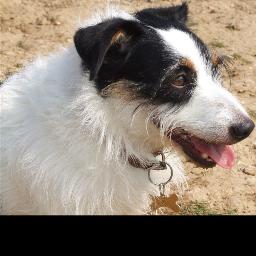}&
\includegraphics[width=.625in]{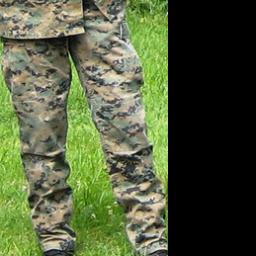}&
\includegraphics[width=.625in]{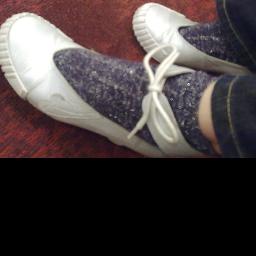}&
\includegraphics[width=.625in]{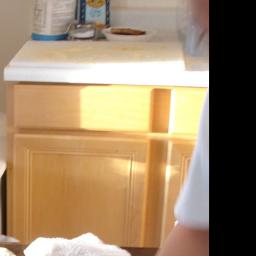}&
\includegraphics[width=.625in]{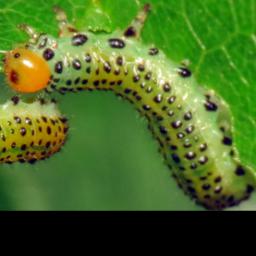}\\
(a) & (b) & (c) & (d) & (e) \\
\end{tabular}
\end{center}
\caption{An example image (top) of a running dog and the lower body of a human.
The image is padded to form a square and downscaled. The composite
embedding with $f$ is computed and then queried with $h$ about the presence of the object in images (a-e),
containing \emph{dog}, \emph{trousers}, \emph{footwear}, \emph{countertop}, and \emph{caterpillar}.
The query function $h$, when given the embeddings of the top image and another image,
should return True for (a,b,c) and False for (d,e).}
\label{fig:OpenImages}
\end{figure}
Here we trained and evaluated Model II on Open Images \cite{OpenImages}.
This dataset contains a total of 16M bounding boxes for 600 object classes on 1.9M images.
This is a highly challenging problem: in the example in Fig.~\ref{fig:OpenImages},
$f$ has to encode a dog, trousers and footwear; then, given completely different images of these classes (and others), 
$h$ has to decide which objects were present in the original image.
In Open Images, each image may contain objects from multiple classes, and each object has a bounding box.
We acquire singleton samples by using the bounding boxes to crop singleton objects from images. 
In this experiment, 500 classes are selected for training and 73 other classes for testing.
The training and evaluation settings are the same as Experiment 2. 

{\bf Architectures}:
For $f$, we use ResNet-18 that was modified to have a 32-dimensional output.
We used the same $h$ as in Experiment 2.

{\bf Baselines}:
\begin{enumerate}
\item {\bf TradEmb}: Similar to Model II, here we compare with a non-compositional embedding trained using one-shot learning
on singleton classes ({\bf TradEm}). All objects are cropped according to their labeled bounding boxes
and then resized and padded to $256\times 256$. All original images are also resized and padded to the same size.

\item {\bf SlideWin}: In 
Open Images, multiple objects co-occur 
in the same image but rarely overlap. Hence, one might wonder how well the following approach would work (somewhat similar to
\cite{hsieh2019one} on one-shot \emph{detection}): train 
a traditional embedding model on examples of cropped objects;  then apply it repeatedly to many ``windows'' within each test image (like a sliding window).
To answer a query about whether the test image contains a certain object, compute the minimum (or median, or other statistic)
distance between the embedding of each window and the embedding of the queried object.

We trained a baseline model using this approach
(accuracy in 2-way forced-choice task on pre-cropped 256x256 images not seen during training: 93.6\%). To answer queries, 
we partitioned each test image into
a rectangular grid of at most 4x4 cells (depending on image aspect ratio).
We then constructed windows corresponding to all possible contiguous subgrids
(there were between 70-100 windows for each image), and then resized each window to 256x256 pixels.
We found that taking the minimum embedding distance
worked best.
\end{enumerate}

\begin{table}
\begin{center}
\setlength{\tabcolsep}{1.6pt}
\begin{tabular}{l||c|c|c||c|c}
\multicolumn{6}{c}{\bf Experiment 3 (Open Images)} \\\hline \hline
       & $h_\textrm{DNN}$ & $h_\textrm{Lin+FC}$ & $h_\textrm{Lin}$ & TradEm & SlideWin  \\ \hline
Acc \% & {\bf 76.9} & 76.8 & 50.0 & 50.1 & 52.6 \\ \hline
AUC    & {\bf 85.4} & 85.2 & 50.3 & 59.2 & 52.1
\end{tabular}
\end{center}
\caption{One-shot learning results for Model II on Open Images compared to either
the TradEm or the SlideWin baselines (similar to \cite{hsieh2019one}).}
\label{fig:query_openimages}
\end{table}

{\bf Results} are shown in Table~\ref{fig:query_openimages}. The compositional models of $f$ combined with 
either $h_\textrm{DNN}$ and $h_\textrm{Lin+FC}$ (though not with $h_\textrm{Lin}$) achieve an AUC of over $85\%$ and easily outperform TradEm.
It also outperforms the SlideWin method: even though this baseline was trained to be 
highly accurate on \emph{pre-cropped} windows (as reported above), it was at-chance when forced to 
aggregate across many windows and answer the containment queries. 
It is also much more slower than the compositional embedding approach.


{\bf Discussion}: An interesting phenomenon we discovered is that while the linear model $g_\textrm{Lin}$ achieves the best results in the $f\& g$ setting
(set union), it is hardly better than random chance for the $f\&h$ setting (set containment).
On the other hand, while $g_\textrm{DNN}$ is worse than other trainable $g$ functions
for set union, it outperforms the other functions for set containment. 
One possible explanation is that
training $f$ in Model I to distinguish explicitly between  all possible subsets causes $f$ to become 
very powerful (relative to the $f$ in Model II), after which only a simple $g$ is needed for set unions. 
The training procedure in Model II based on set containment might provide less information to $f$, thus requiring $g$ to be more
powerful to compensate.
Another possibility is that, since $g$ is applied recursively to construct  unions, its complexity
must be kept small to avoid overfitting.

\section{Model III (supervised): $f_\textrm{im}$, $f_\textrm{label}$, \& $h$}
Given the promising results on one-shot learning tasks for object recognition, we wanted 
to assess whether compositional embeddings could be beneficial for multi-label classification
in standard supervised  learning problems where the testing and training classes are the same (i.e., \emph{not} one-shot).
Specifically, we developed a model to answer questions of the form, ``Does image $x$ contain an object of class $y$?''.
The intuition is that a compositional embedding approach might make the recognition process more accurate by giving it
knowledge of \emph{which} object is being queried \emph{before} analyzing the input image for its contents.
Model III consists of \emph{three} functions trained jointly:
(1) a deep embedding $f_\textrm{im}$ for the input image $x$; (2) a linear layer
$f_\textrm{label}$ to embed a one-hot vector of the queried label $y$ into a distributed representation;
and (3) a query function $h$ that takes the two embeddings
as inputs and outputs the probability that the image contains an object with the desired label. 
This approach enables the combined model to modulate its perception of the objects contained within an image based
on the specific task, i.e., the specific label that was queried, which may help it to perform more accurately \cite{mozer2008top}.

\subsection{Experiment 4: COCO}
\begin{figure}
\begin{center}
\includegraphics[width=\columnwidth]{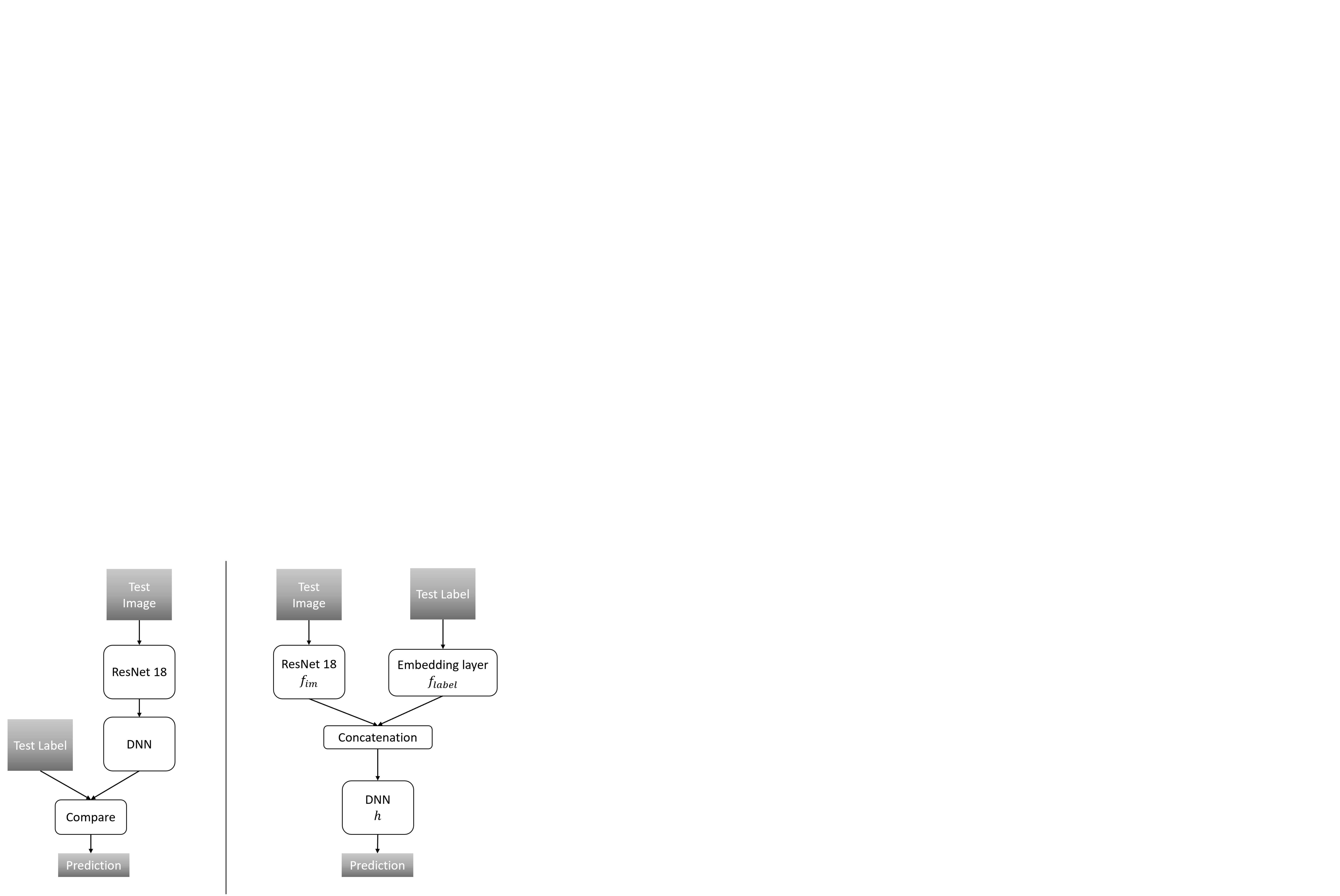}
\end{center}
\caption{{\bf Supervised multi-label image classification}: Left is a traditional approach based on a CNN with
multiple independent sigmoid outputs. Right is the proposed Model III with 3 jointly trained embeddings
$f_\textrm{im}$, $f_\textrm{label}$, \& $h$.}
\label{fig:COCO}
\end{figure}
To evaluate Model III, we conducted an
experiment on the Microsoft COCO dataset \cite{lin2014microsoft},
which has $|\mathcal{S}|=80$ classes in both the training and validation sets.
During evaluation, half the test labels are positive and the other half are negative.

{\bf Architecture}: For $f_\textrm{im}$, we modify ResNet-18 (pretrained on ImageNet) so
that its last layer has dimension 128. The embedding layer $f_\textrm{label}$ maps
80-dimensional 1-hot labels to 32-dimension real-valued embeddings. Then the image embedding and label embedding are concatenated to a 160-dimension vector
and fed to the
DNN $h$, consisting of
$\textrm{FC}(160)\rightarrow \textrm{BN}\rightarrow \textrm{ReLU}\rightarrow \textrm{FC}(136)\rightarrow \textrm{BN}\rightarrow \textrm{ReLU}\rightarrow \textrm{FC}(136)\rightarrow \textrm{Sigmoid}(1)$,
where $\textrm{Sigmoid}(k)$ is a sigmoidal layer with $k$ independent probabilistic outputs. The output of $h$ represents the probability that
image $x$ contains an object of class $y$.
See Figure \ref{fig:COCO} (right). 
Binary cross-entropy, summed over all classes, is used as the loss function.
Because of class imbalance, different weights are used for positive and negative classes according to their numbers in each image.

{\bf Baseline}:  We compare to a baseline consisting of a pretrained ResNet-18 followed by a DNN (to enable a fairer comparison with our Model III).
The DNN consists of
$\textrm{FC}(128)\rightarrow \textrm{BN}\rightarrow \textrm{ReLU}\rightarrow \textrm{FC}(128)\rightarrow \textrm{BN}\rightarrow \textrm{ReLU}\rightarrow \textrm{FC}(128)\rightarrow \textrm{Sigmoid}(80)$. The final layer gives independent probabilistic predictions of the 80 classes.
Note that this DNN has almost exactly the same number of parameters as the DNN for Model III.
For multi-label image classification, we simply check whether the output for the desired label is close to 1.
See Figure \ref{fig:COCO} (left).

{\bf Results}: The baseline accuracy using the ResNet attained an accuracy 
of $64.0\%$ and AUC of $67.7\%$. In contrast, the compositional embedding approach ($f_\textrm{im}\&f_\textrm{label}\&h$) achieved
a substantially higher accuracy of $82.0\%$ and AUC is $90.8\%$. This accuracy improvement may stem from the task modulation
of the visual processing, or from the fact that the compositional method was explicitly designed to answer binary image
queries rather than represent the image as a $|\mathcal{S}|$-dimensional vector (as with a standard object recognition CNN).

\section{Conclusions}
We developed a compositional embedding mechanism whereby the \emph{set} of objects contained in the
input data must be both \emph{perceived} and then mapped into a space such that the
\emph{set relationships} -- union (Model I) and containment (Model II) -- between multiple embedded vectors can be inferred.
Importantly, the ground-truth rendering process for how examples are rendered from their component classes 
must implicitly be learned.
This new domain of \emph{multi-label one-shot learning} is highly challenging but has interesting applications to multi-object
image recognition in computer vision, as well as multi-person speaker recognition and diarization in computer audition.
In contrast to prior work \cite{alfassy2019laso,huynh2020shared}, our models require only relatively weak one-shot supervision
consisting of the label-set relationships among the training examples.
Our experiments on OmniGlot, Open Images, and COCO show promising results: the compositional embeddings
strongly outperformed baselines based on traditional embeddings.  These results provide further evidence that embedding functions can encode
rich and complex structure about the \emph{multiple} objects contained in the images they came from.
Our results also shed light on how the task structure
influences the best design of the functions $f$, $g$, and $h$. Finally, we demonstrated the potential of compositional
embeddings for standard supervised tasks of multi-label image recognition (Model III): task-specific perception of images,
as enabled by jointly trained embedding functions, can boost perceptual accuracy.

One direction for {\bf future research} -- motivated by perceptual expertise research on, for example, how
chess experts perceive real vs.~random game configurations \cite{chase1973perception} -- 
is to take better advantage 
of the class co-occurrence structure in a specific application domain (e.g., which objects co-occur in images).

\section*{Acknowledgements}
This material is based on work supported by the National Science Foundation Cyberlearning grant \#1822768.

{\small
\bibliographystyle{ieee_fullname}
\bibliography{reference}
}

\end{document}